# Web Usage Mining Using Artificial Ant Colony Clustering and Genetic Programming

**Ajith Abraham**
Department of Computer Science,
Oklahoma State University, Tulsa, OK 74106, USA
aa@cs.okstate.edu

**Vitorino Ramos**
CVRM-GeoSystems Centre,
Technical University of Lisbon, Portugal
vitorino.ramos@alfa.ist.utl.pt

**Abstract-** The rapid ecommerce growth has made both business community and customers face a new situation. Due to intense competition on one hand and the customer's option to choose from several alternatives business community has realized the necessity of intelligent marketing strategies and relationship management. Web usage mining attempts to discover useful knowledge from the secondary data obtained from the interactions of the users with the Web. Web usage mining has become very critical for effective Web site management, creating adaptive Web sites, business and support services, personalization, network traffic flow analysis and so on. The study of ant colonies behavior and their self-organizing capabilities is of interest to knowledge retrieval/ management and decision support systems sciences, because it provides models of distributed adaptive organization, which are useful to solve difficult optimization, classification, and distributed control problems, among others [17][18][16]. In this paper, we propose an ant clustering algorithm to discover Web usage patterns (data clusters) and a linear genetic programming approach to analyze the visitor trends. Empirical results clearly shows that ant colony clustering performs well when compared to a self-organizing map (for clustering Web usage patterns) even though the performance accuracy is not that efficient when comparared to evolutionary-fuzzy clustering (*i-miner*) [1] approach.

## 1 Introduction

The WWW continues to grow at an amazing rate as an information gateway and as a medium for conducting business. Web mining is the extraction of interesting and useful knowledge and implicit information from atrifacts or activity related to the WWW [12][7]. Web servers record and accumulate data about user interactions whenever requests for resources are received. Analyzing the Web access logs can help understand the user behaviour and the web structure. From the business and applications point of view, knowledge obtained from the Web usage patterns could be directly applied to efficiently manage activities related to e-business, eservices, e-education and so on. Accurate Web usage information could help to attract new customers, retain current customers, improve cross marketing/sales, effectiveness of promotional campaigns, tracking leaving customers and find the most effective logical structure for their Web space. User profiles could be built by combining users' navigation paths with other data features, such as page viewing time, hyperlink structure, and page content [9]. What makes the discovered knowledge interesting had been addressed by several works. Results previously known are very often considered as not interesting. So the key concept to make the discovered knowledge interesting will be its novelty or unexpectedness appearance.

There are several commercial softwares that could provide Web usage statistics. These stats could be useful for Web administrators to get a sense of the actual load on the server. For small web servers, the usage statistics provided by conventional Web site trackers may be adequate to analyze the usage pattern and trends. However as the size and complexity of the data increases, the statistics provided by existing Web log file analysis tools may prove inadequate and more intelligent mining techniques will be necessary [10].

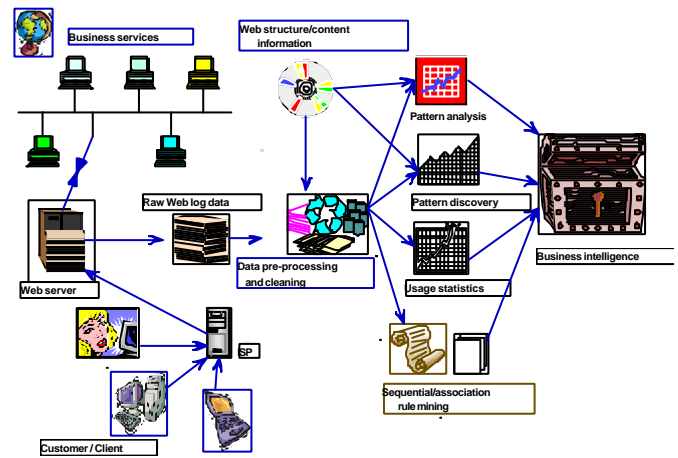

**Figure 1.** Web usage mining framework

A generic Web usage mining framework is depicted in Figure 1. In the case of Web mining data could be collected at the server level, client level, proxy level or some consolidated data. These data could differ in terms of content and the way it is collected etc. The usage data collected at different sources represent the navigation patterns of different segments of the overall Web traffic, ranging from single user, single site browsing behaviour to multi-user, multi-site access patterns. As evident from Figure 1, Web server log does not accurately contain sufficient information for infering the behaviour at the client side as they relate to the pages served by the Web server. Pre-procesed and cleaned data could be used for pattern discovery, pattern analysis, Web usage statistics and generating association/ sequential rules. Much work has been performed on extracting various pattern information from Web logs and the application of the discovered knowledge range from improving the design and structure of a Web site to enabling business organizations to function more effeciently. Jespersen et al [10] proposed an hybrid approach for analyzing the visitor click sequences. A combination of hypertext probabilistic grammar and click fact table approach is used to mine Web logs which could be also used for general sequence mining tasks. Mobasher et al [14] proposed the Web personalization system which consists of offline tasks related to the mining of usage data and online process of automatic Web page customization based on the knowledge discovered. LOGSOM proposed by Smith et al [19], utilizes self-organizing map to organize web pages into a two-dimensional map based solely on the users' navigation behavior, rather than the content of the web pages. LumberJack proposed by Chi et al [6] builds up user profiles by combining both user session clustering and traditional statistical traffic analysis using K-means algorithm. Joshi et al [11] used relational online analytical processing approach for creating a Web log warehouse using access logs and mined logs (association rules and clusters). A comprehensive overview of Web usage mining research is found in [7][20].

In this paper, an ant colony clustering (ACLUSTER) [16] is proposed to seggregate visitors and thereafter a linear genetic programming approach [3] to analyze the visitor trends. The results are compared with the earlier works using self organizing map [21] and evolutionary - fuzzy c means algorithm [1] to seggregate the user access records and several soft computing paradigms to analyze the user access trends.

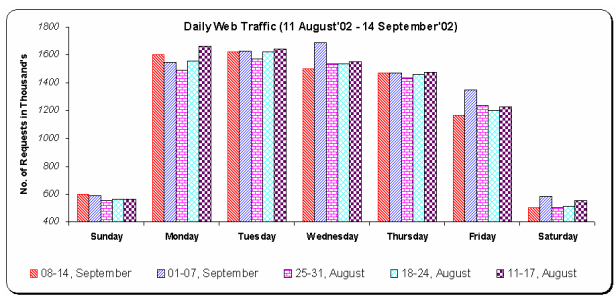

**Figure 2.** University's daily Web traffic pattern for 5 weeks [15]

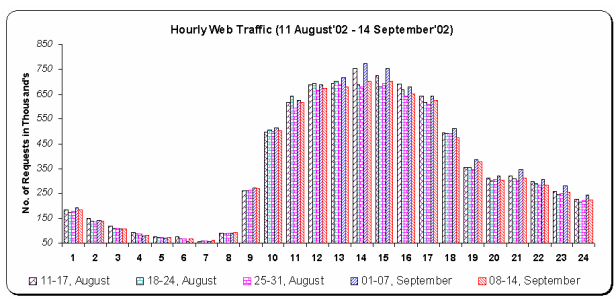

**Figure 3.** Average hourly Web traffic patterns for 5 weeks [15]

Web access log data at the Monash University's Web site [15] were used for experimentations. The University's central web server receives over 7 million hits in a week and therefore it is a real challenge to find and extract hidden usage pattern information. Average daily and hourly access patterns for 5 weeks (11 August'02 – 14 September'02) are shown in Figures 2 and 3 respectively. The average daily and hourly patterns even though tend to follow a similar trend (as evident from the figures) the differences tend to increase during high traffic days (Monday – Friday) and during the peak hours (11:00-17:00 Hrs). Due to the enormous traffic volume and chaotic access behavior, the prediction of the user access patterns becomes more difficult and complex.

In the subsequent section, we present the proposed architecture and experimentation results of the ant clustering – linear genetic programming approach. Some conclusions are provided towards the end.

## 2. Hybrid Framework Using Ant Colony Clustering and Linear Genetic Programming Approach (ANT-LGP)

The hybrid framework uses an ant colony optimization algorithm to cluster Web usage patterns. The raw data from the log files are cleaned and pre-processed and the ACLUSTER algorithm [16] is used to identify the usage patterns (data clusters). The developed clusters of data are fed to a linear genetic programming model to analyze the usage trends.

### 2.1 Ant Colony Clustering Using Bio-Inspired Spatial Probabilities

In several species of ants, workers have been reported to sort their larvae or form piles of corpses – literally cemeteries – to clean up their nests. Chrétien [4] has performed experiments with the ant *Lasius niger* to study the organization of cemeteries. Other experiments include the ants *Pheidole pallidula* reported by Deneubourg et al. [8]. In Nature many species actually organize a cemetery. If corpses, or more precisely, sufficiently large parts of corpses are randomly distributed in space at the beginning of the experiment, the workers form cemetery clusters within a few hours, following a behavior similar to aggregation. If the experimental arena is not sufficiently large, or if it contains spatial heterogeneities, the clusters will be formed along the edges of the arena or, more generally, following the heterogeneities. The basic mechanism underlying this type of aggregation phenomenon is an attraction between dead items mediated by the ant workers: small clusters of items grow by attracting workers to deposit more items. It is this positive and auto-catalytic feedback that leads to the formation of larger an larger clusters. In this case, it is therefore the distribution of the clusters in the environment that plays the role of stigmergic variable. Denebourg et al. [8] have proposed one model (BM -basic model) to account for the above-mentioned phenomenon of corpse clustering in ants. The general idea is that isolated items should be picked up and dropped at some other location where more items of that type are present. Lumer's and Faieta (LF) model [13] have generalized Denebourg et al.'s basic method [8], and applied it to exploratory data analysis.

Instead of trying to solve some disparities in the basic LF algorithm by adding different ant casts, short-term memories and behavioral switches, which are computationally intensive, representing simultaneously a potential and difficult complex parameter tuning, Ramos et al [16] proposed ACLUSTER algorithm to follow real ant-like behaviors as much as possible. In that sense, bio-inspired spatial transition probabilities are incorporated into the system, avoiding randomly moving agents, which tend the distributed algorithm to explore regions manifestly without interest (e.g., regions without any type of object clusters), being generally, this type of exploration, counterproductive and time consuming. Since this type of transition probabilities depends on the spatial distribution of pheromone across the environment, the behavior reproduced is also a stigmergic one. Moreover, the strategy not only allows guiding ants to find clusters of objects in an adaptive way (if, by any reason, one cluster disappears, pheromone tends to evaporate on that location), as the use of embodied short-term memories is avoided (since this transition probabilities tends also to increase pheromone in specific locations, where more objects are present). As we shall see, the distribution of the pheromone represents the memory of the recent history of the swarm, and in a sense it contains information, which the individual ants are unable to hold or transmit. There is no direct communication between the organisms but a type of indirect communication through the pheromonal field. In fact, ants are not allowed to have any memory and the individual's spatial knowledge is restricted to local information about the whole colony pheromone density.

In order to model the behavior of ants associated to different tasks, as dropping and picking up objects, we suggest the use of combinations of different response thresholds. As we have seen before, there are two major factors that should influence any local action taken by the ant-like agent: the number of objects in his neighborhood, and their similarity (including the hypothetical object carried by one ant). Lumer and Faieta [13] use an average similarity, mixing distances between objects with their number, incorporating it simultaneously into a response threshold function like the one of Denebourg's [8]. Instead, ACLUSTER uses combinations of two independent response threshold functions, each associated with a different environmental factor (or, stimuli intensity), that is, the number of objects in the area, and their similarity. The computation of average similarities are avoided in the ACLUSTER algorithm, since this strategy could be somehow blind to the number of objects present in one specific neighborhood. Bonabeau et al. [5], proposed a family of response threshold functions in order to model response thresholds. Every individual has a response threshold for every task. Individuals engage in task performance when the level of the task-associated stimuli $s$, exceeds their thresholds. Technical details of ACLUSTER could be obtained from [16]

### 2.2 Experimentation Setup and Clustering Results

In this research, we used the statistical/ text data generated by the log file analyzer from 01 January 2002 to 07 July 2002. Selecting useful data is an important task in the data pre-processing block. After some preliminary analysis, we selected the statistical data comprising of domain byte requests, hourly page requests and daily page requests as focus of the cluster models for finding Web users' usage patterns. The most recently accessed data were indexed higher while the least recently accessed data were placed at the bottom.

For each of datasets (daily and hourly Web log data ), the algorithm was run twice (for $t$= 1 to 10,00,000) in order to check if somehow the results were similar (which appear to be, if we look onto which data items are connected into what clusters). The classification space is always 2D, non-parametric and toroidal. Experimentation results for the daily and hourly Web traffic data are presented in Figures 4 and 5.

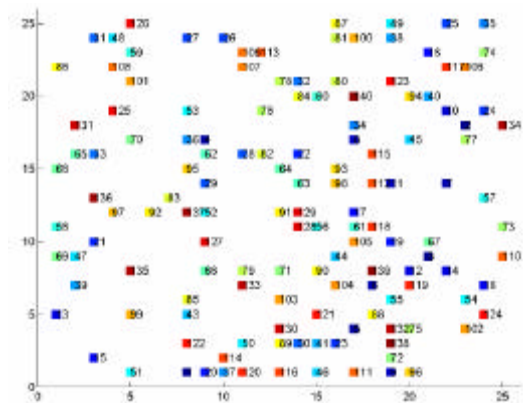

**(a)** t=1

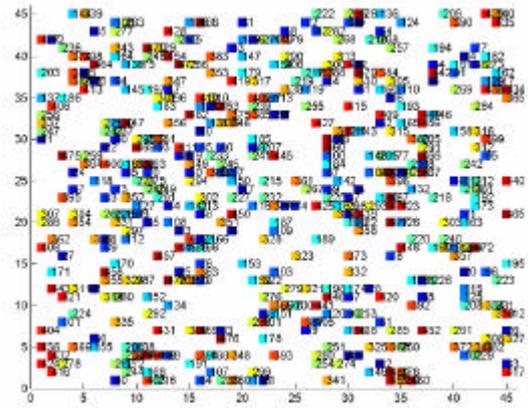

(a) t=1

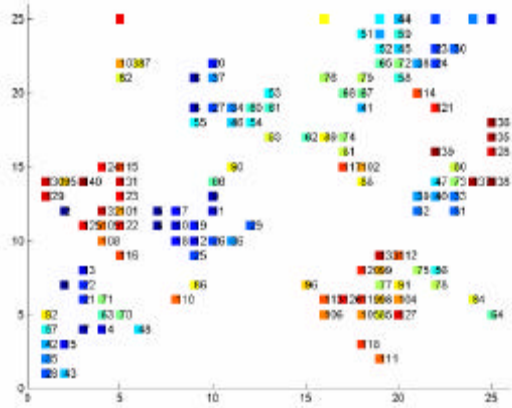

(b) t= 100

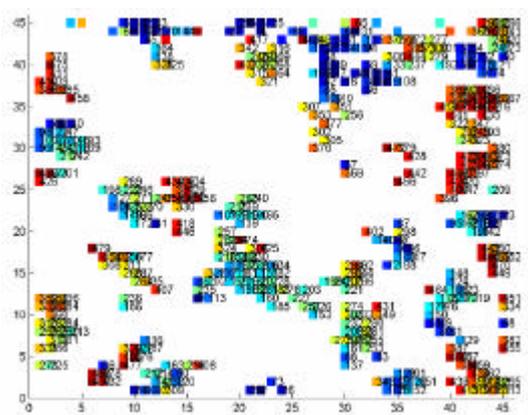

(b) t= 100

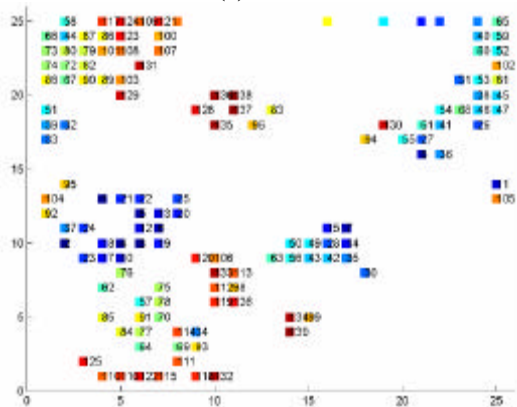

(c) t= 500

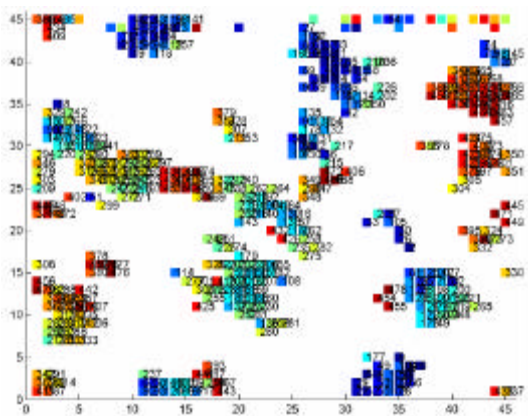

(c) t= 500

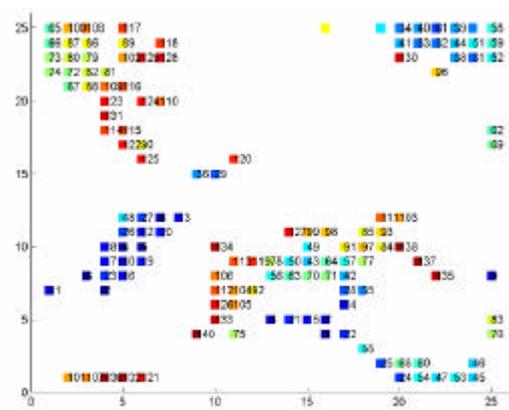

(d) t= 10,00,000

**Figure 4 (a-d).** The snapshots represent the spatial distribution of daily Web traffic data on a 25 x 25 non-parametric toroidal grid at several time steps. At *t*=1, data items are randomly allocated into the grid. As time evolves, several homogenous clusters emerge due to the ant colony action. Type 1 probability function was used with $k_1$= 0.1, $k_2$= 0.3 and 14 ants. ($k_1$ and $k_2$ are threshold constants)

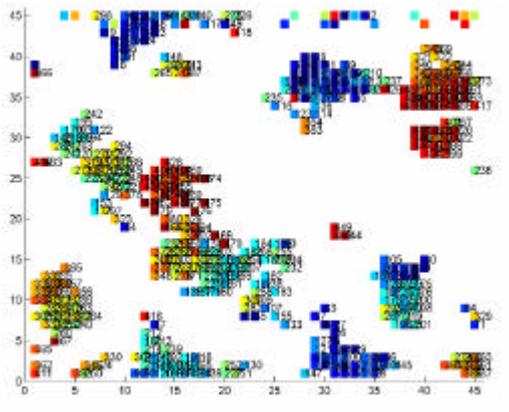

**(d)** t= 10,00,000

**Figure 5 (a-d).** The snapshots represent the spatial distribution of hourly Web traffic data on a 45 x 45 non-parametric toroidal grid at several time steps. At $t=1$, data items are randomly allocated into the grid. As time evolves, several homogenous clusters emerge due to the ant colony action. Type 1 probability function was used with $k_1= 0.1$, $k_2= 0.3$ and 48 ants.

### 2.3 Linear Genetic Programming (LGP)

Linear genetic programming is a variant of the GP technique that acts on linear genomes [3]. Its main characteristics in comparison to tree-based GP lies in that the evolvable units are not the expressions of a functional programming language (like LISP), but the programs of an imperative language (like c/c ++). An alternate approach is to evolve a computer program at the machine code level, using lower level representations for the individuals. This can tremendously hasten up the evolution process as, no matter how an individual is initially represented, finally it always has to be represented as a piece of machine code, as fitness evaluation requires physical execution of the individuals.

The basic unit of evolution here is a native machine code instruction that runs on the floating-point processor unit (FPU). Since different instructions may have different sizes, here instructions are clubbed up together to form instruction blocks of 32 bits each. The instruction blocks hold one or more native machine code instructions, depending on the sizes of the instructions. A crossover point can occur only between instructions and is prohibited from occurring within an instruction. However the mutation operation does not have any such restriction.

### 2.4 Experimentation Setup and Trend Analysis Results

Besides the inputs '*volume of requests*' and '*volume of pages (bytes)*' and '*index number*', we also used the '*cluster information*' provided by the clustering algorithm as an additional input variable. The data was re-indexed based on the cluster information. Our task is to predict (few time steps ahead) the Web traffic volume on a hourly and daily basis. We used the data from 17 February 2002 to 30 June 2002 for training and the data from 01 July 2002 to 06 July 2002 for testing and validation purposes.

We used a LGP technique that manipulates and evolves program at the machine code level. We used the Discipulus workbench for simulating LGP [[2]]. The settings of various linear genetic programming system parameters are of utmost importance for successful performance of the system. The population space has been subdivided into multiple subpopulation or demes. Migration of individuals among the subpopulations causes evolution of the entire population. It helps to maintain diversity in the population, as migration is restricted among the demes. Moreover, the tendency towards a bad local minimum in one deme can be countered by other demes with better search directions. The various LGP search parameters are the mutation frequency, crossover frequency and the reproduction frequency: The crossover operator acts by exchanging sequences of instructions between two tournament winners. After a trial and error approach, the following parameter settings were used for the experiments.

Population size: 500
Maximum no. of tournaments : 120,000
Mutation frequency: 90%
Crossover frequency: 80%
Number of demes: 10
Maximum program size: 512
Target subset size: 100

The experiments were repeated three times and the test data was passed through the saved model. Figures 6 and 8 illustrate the average growth in program length for hourly and daily Web traffic. Figures 7 and 9 depict the training and test performance (average training and test fitness) for hourly and daily Web traffic. Empirical comparison of the proposed framework with some of our previous work is depicted in Tables 1 and 2. Performance comparison of the proposed framework (ANT-LGP) with *i-Miner* (hybrid evolutionary fuzzy clustering–fuzzy inference system) [1], self-organizing map–linear genetic programming (SOM-LGP) [21] and self-organizing map–artificial neural network (SOM-ANN) [21] are graphically illustrated in Figures 10 and 11.

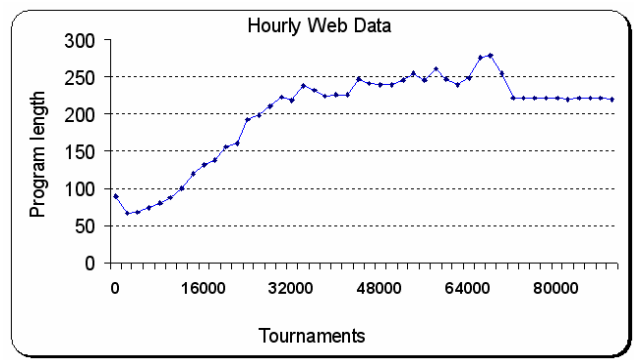

**Figure 6.** Hourly Web data analysis: Growth in average program length during 120,000 tournaments.

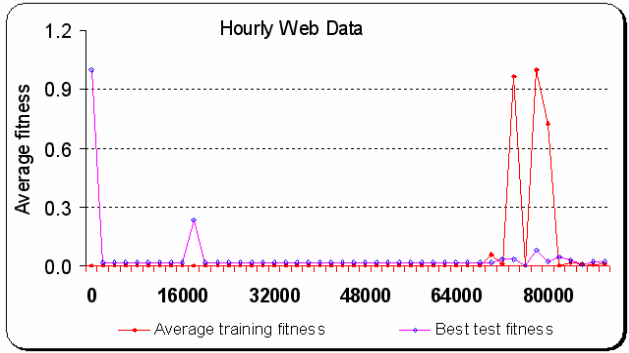

**Figure 7.** Hourly Web data analysis: Comparison of average training and test fitness during 120,000 tournaments.

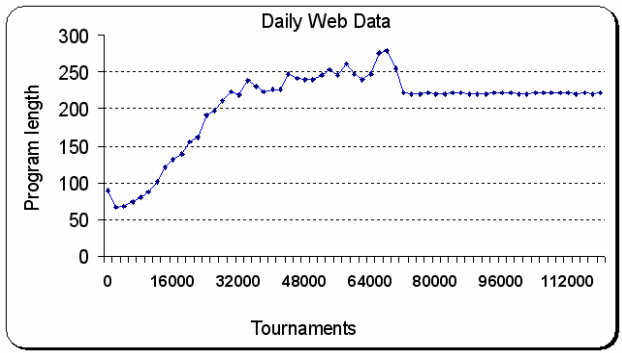

**Figure 8.** Daily Web data analysis: growth in average program length during 120,000 tournaments.

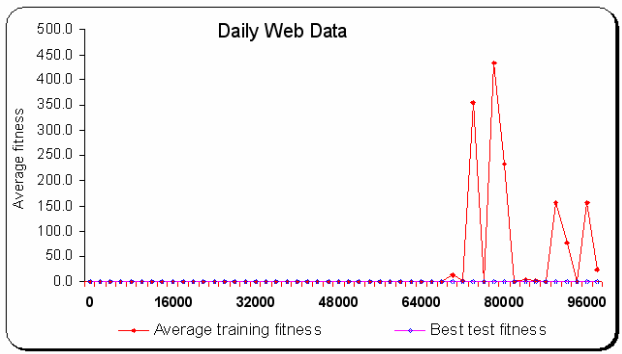

**Figure 9.** Daily Web data analysis: comparison of average training and test fitness during 120,000 tournaments.

**Table 1.** Performance of the different paradigms for daily Web data

| Method | Daily (1 day ahead) | | CC |
|---|---|---|---|
| | RMSE | | |
| | Train | Test | |
| ANT-LGP | 0.0191 | 0.0291 | 0.9963 |
| i-Miner[#] | 0.0044 | 0.0053 | 0.9967 |
| ANN[*] | 0.0345 | 0.0481 | 0.9292 |
| LGP[*] | 0.0543 | 0.0749 | 0.9315 |

**Table 2.** Performance of the different paradigms for hourly Web data

| Method | Hourly (1 hour ahead) | | CC |
|---|---|---|---|
| | RMSE | | |
| | Train | Test | |
| ANT-LGP | 0.2561 | 0.035 | 0.9921 |
| i-Miner- (FCM-FIS) | 0.0012 | 0.0041 | 0.9981 |
| SOM-ANN | 0.0546 | 0.0639 | 0.9493 |
| SOM-LGP | 0.0654 | 0.0516 | 0.9446 |

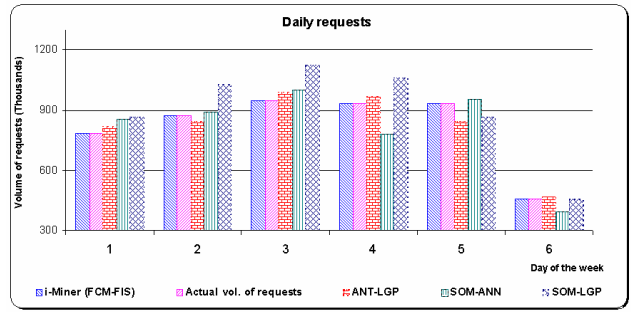

**Figure 10.** Comparison of different paradigms for daily Web traffic trends

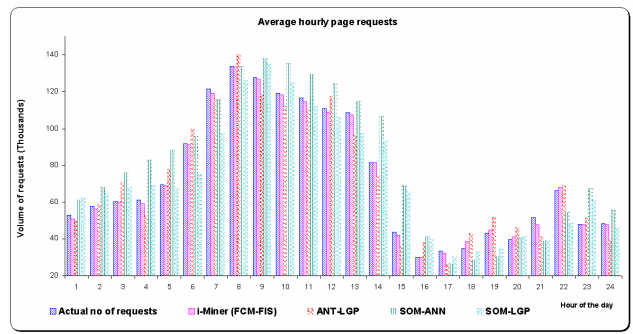

**Figure 11.** Comparison of different paradigms for hourly Web traffic trends

## 3. Conclusions

The proposed ANT-LGP model seems to work very well for the problem considered. The empirical results also reveal the importance of using optimization techniques for mining useful information. In this paper, our focus was to develop accurate trend prediction models to analyze the hourly and daily web traffic volume. Several useful information could be discovered from the clustered data. The knowledge discovered from the developed clusters using different intelligent models could be a good comparison study and is left as a future research topic.

As illustrated in Tables 1 and 2, incorporation of the ant clustering algorithm helped to improve the performance of the LGP model (when compared to clustering using self organizing maps). *i-Miner* framework gave the overall best results with the lowest RMSE on test error and the highest correlation coefficient.

Future research will also incorporate more data mining algorithms to improve knowledge discovery and association rules from the clustered data. The contribution of the individual input variables and different clustering algorithms will be also investigated to improve the trend analysis and knowledge discovery.

## Bibliography

[1] Abraham A., i-Miner: A Web Usage Mining Framework Using Hierarchical Intelligent Systems, The IEEE International Conference on Fuzzy Systems FUZZ-IEEE'03, pp. 1129-1134, 2003.

[2] AIMLearning Technology, http://www.aimlearning.com

[3] Banzhaf. W., Nordin. P., Keller. E. R., Francone F. D. (1998) "Genetic Programming : An Introduction on The Automatic Evolution of Computer Programs and its Applications," Morgan Kaufmann Publishers, Inc.

[4] Bonabeau E., M. Dorigo, G. Théraulaz, Swarm Intelligence: From Natural to Artificial Systems. Santa Fe Institute in the Sciences of the Complexity, Oxford University Press, New York, Oxford, 1999.

[5] Bonabeau, E., Théraulaz, G., Denebourg, J.-L., "Quantitative Study of the Fixed Response Threshold Model for the Regulation of Division of Labour in Insect Societies". *Roy. Soc. B*, 263, pp.1565-1569, 1996.

[6] Chi E.H., Rosien A. and Heer J., LumberJack: Intelligent Discovery and Analysis of Web User Traffic Composition. In Proceedings of ACM-SIGKDD Workshop on Web Mining for Usage Patterns and User Profiles, Canada, pp.. , ACM Press, 2002.

[7] Cooley R., Web Usage Mining: Discovery and Application of Interesting patterns from Web Data, Ph. D. Thesis, Department of Computer Science, University of Minnesota, 2000.

[8] Deneubourg, J.-L., Goss, S., Franks, N., Sendova-Franks A., Detrain, C., Chretien, L. "The Dynamic of Collective Sorting Robot-like Ants and Ant-like Robots", SAB'90 - 1st Conf. On Simulation of Adaptive Behavior: From Animals to Animats, J.A. Meyer and S.W. Wilson (Eds.), 356-365. MIT Press, 1991.

[9] Heer, J. and Chi E.H., Identification of Web User Traffic Composition using Multi- Modal Clustering and Information Scent, In *Proc. of the Workshop on Web Mining,* SIAM Conference on Data Mining, pp. 51-58, 2001

[10] Jespersen S.E., Thorhauge J., and Bach T., A Hybrid Approach to Web Usage Mining, Data Warehousing and Knowledge Discovery, LNCS 2454, Y. Kambayashi, W. Winiwarter, M. Arikawa (Eds.), pp. 73-82, 2002.

[11] Joshi, K.P., Joshi, A., Yesha, Y., Krishnapuram, R., Warehousing and Mining Web Logs. Proceedings of the 2nd ACM CIKM Workshop on Web Information and Data Management, pp. 63-68, 1999.

[12] Kosala R and Blockeel H., Web Mining Research: A Survey, ACM SIGKDD Explorations, 2(1), pp. 1-15, 2000.

[13] Lumer E. D. & Faieta B., Diversity and Adaptation in Populations of Clustering Ants. In Cliff, D., Husbands, P., Meyer, J. and Wilson S. (Eds.), *in* From Animals to Animats 3, Proc. of the 3rd Int. Conf. on the Simulation of Adaptive Behavior. Cambridge, MA: The MIT Press/Bradford Books, 1994.

[14] Mobasher B., Cooley R. and Srivastava J., Creating Adaptive Web Sites through Usage-based Clustering of URLs, In Proceedings of 1999 Workshop on Knowledge and Data Engineering Exchange, USA, pp.19-25, 1999.

[15] Monash University Web site: http://www.monash.edu.au

[16] Ramos Vitorino, Fernando Muge, Pedro Pina, Self-Organized Data and Image Retrieval as a Consequence of Inter-Dynamic Synergistic Relationships in Artificial Ant Colonies, Soft Computing Systems - Design, Management and Applications, 2nd Int. Conf. on Hybrid Intelligent Systems, IOS Press, pp. 500-509, 2002.

[17] Ramos, Vitorino and Merelo, Juan J. "Self-Organized Stigmergic Document Maps: Environment as a Mechanism for Context Learning", in E. Alba, F. Herrera, J.J. Merelo et al. (Eds.), AEB´02 - 1st Int. Conf. On Metaheuristics, Evolutionary and Bio-Inspired Algorithms, pp. 284-293, Mérida, Spain, 2002.

[18] Ramos, Vitorino, and Almeida, F. "Artificial Ant Colonies in Digital Image Habitats - A Mass Behaviour Effect Study on Pattern Recognition", in Marco Dorigo, Martin Middendorf and Thomas Stüzle (Eds.), Proc. Of ANTS'00 – 2nd Int. Workshop on Ant Algorithms, pp. 113-116, Brussels, Belgium, 2000.

[19] Smith K.A. and Ng A., Web page clustering using a self-organizing map of user navigation

patterns,Decision Support Systems, Volume 35, Issue 2 , pp. 245-256, 2003.

[20] Srivastava, J., Cooley R., Deshpande, M., Tan, P.N., Web Usage Mining: Discovery and Applications of Usage Patterns from Web Data. SIGKDD Explorations, vol. 1, no. 2, pp. 12-23, 2000.

[21] Wang X., Abraham A. and Smith K.A, Soft Computing Paradigms for Web Access Pattern Analysis, Proceedings of the 1st International Conference on Fuzzy Systems and Knowledge Discovery, pp. 631-635, 2002.